\documentclass[conference]{IEEEtran}
\usepackage[utf8]{inputenc} % allow utf-8 input
\usepackage[T1]{fontenc}    % use 8-bit T1 fonts
\usepackage{hyperref}       % hyperlinks
\usepackage{url}            % simple URL typesetting
\usepackage{booktabs}       % professional-quality tables
\usepackage{amsfonts}       % blackboard math symbols
\usepackage{nicefrac}       % compact symbols for 1/2, etc.
\usepackage{microtype}      % microtypography
\usepackage{xcolor}         % colors
\usepackage{amsmath,graphicx}
\usepackage{float}
\usepackage{tabularx}

\usepackage{graphicx}
\usepackage{xcolor}
\usepackage{tabu}
\usepackage{multirow}
\usepackage{cite}
\usepackage{amsmath}
\usepackage{amssymb}
\usepackage{amsthm}
\usepackage{amsfonts}
\usepackage{indentfirst}
\usepackage{tabto}
\usepackage{pifont}
\usepackage{epstopdf}
\usepackage{float}
\usepackage{algorithm}
\usepackage{algorithmic}
\usepackage[utf8]{inputenc}
\usepackage[T1]{fontenc}
\usepackage{cite}
\usepackage{amsmath,amssymb,amsfonts}
\usepackage{algorithmic}
\usepackage{graphicx}
\usepackage{textcomp}
\usepackage{xcolor}
\usepackage{scalerel,xparse}
\usepackage{amsmath,graphicx}
\usepackage{color, colortbl}
\usepackage{float}
\usepackage{tabularx}
\usepackage{booktabs}
\usepackage{url}
\usepackage{paralist}
\usepackage{mathtools}
\usepackage{dblfloatfix}
\usepackage{caption}
\usepackage{orcidlink}

\definecolor{Gray}{gray}{0.9}
\title{PrivPAS: A real time Privacy-Preserving AI System and applied ethics}

\def\BibTeX{{\rm B\kern-.05em{\sc i\kern-.025em b}\kern-.08em
		T\kern-.1667em\lower.7ex\hbox{E}\kern-.125emX}}
\begin{document}
	
	\title{PrivPAS: A real time Privacy-Preserving AI System and applied ethics}

	\author{
		
		\IEEEauthorblockN{Harichandana B S S}  
		\IEEEauthorblockA{\textit{Samsung R\&D Institute} \\
			Bangalore, India \\
			hari.ss@samsung.com \\
			\orcidlink{0000-0002-6123-2249} 0000-0002-6123-2249
		}\\
		
		\IEEEauthorblockN{Gopi Ramena } 
		
		\IEEEauthorblockA{\textit{Samsung R\&D Institute} \\
			Bangalore, India \\
			gopi.ramena@samsung.com \\
			\orcidlink{0000-0002-7498-3973} 0000-0002-7498-3973
		}
		
		\and
		
		\IEEEauthorblockN{Vibhav Agarwal} 
		
		\IEEEauthorblockA{\textit{Samsung R\&D Institute} \\
			Bangalore, India \\
			vibhav.a@samsung.com \\
			\orcidlink{0000-0002-2029-9885} 0000-0002-2029-9885
		}\\
		
		\IEEEauthorblockN{Sumit Kumar} 
	
		\IEEEauthorblockA{\textit{Samsung R\&D Institute} \\
			Bangalore, India \\
			sumit.kr@samsung.com \\
			\orcidlink{0000-0003-0373-1397} 0000-0003-0373-1397
		}
		
		\and
		\IEEEauthorblockN{Sourav Ghosh} 
		
		\IEEEauthorblockA{\textit{Samsung R\&D Institute} \\
			Bangalore, India \\
			sourav.ghosh@samsung.com \\
			\orcidlink{0000-0003-1866-1408} 0000-0003-1866-1408
		}\\
		
		\IEEEauthorblockN{Barath Raj Kandur Raja}
		\IEEEauthorblockA{\textit{Samsung R\&D Institute} \\
			Bangalore, India \\
			barathraj.kr@samsung.com \\
		\orcidlink{0000-0003-0451-2452}	0000-0003-0451-2452
		}	
	}
	
	\maketitle
	
	\begin{abstract}
		With 3.78 billion social media users worldwide in 2021 (48\% of the human population), almost 3 billion images are shared daily. At the same time, a consistent evolution of smartphone cameras has led to a photography explosion with 85\% of all new pictures being captured using smartphones. However, lately, there has been an increased discussion of privacy concerns when a person being photographed is unaware of the picture being taken or has reservations about the same being shared. These privacy violations are amplified for people with disabilities, who may find it challenging to raise dissent even if they are aware. Such unauthorized image captures may also be misused to gain sympathy by third-party organizations, leading to a privacy breach. Privacy for people with disabilities has so far received comparatively less attention from the AI community. This motivates us to work towards a solution to generate privacy-conscious cues for raising awareness in smartphone users of any sensitivity in their viewfinder content. To this end, we introduce PrivPAS (A real time Privacy-Preserving AI System) a novel framework to identify sensitive content. Additionally, we curate and annotate a dataset to identify and localize accessibility markers and classify whether an image is sensitive to a featured subject with a disability. We demonstrate that the proposed lightweight architecture, with a memory footprint of a mere 8.49MB, achieves a high mAP of 89.52\% on resource-constrained devices. Furthermore, our pipeline, trained on face anonymized data, achieves an F1-score of 73.1\%.  
	\end{abstract}
	
	\begin{IEEEkeywords}
		Accessibility, Privacy, Object Detection, Eye Gaze Detection, Face anonymization
	\end{IEEEkeywords}
	
	\section{Introduction}\label{sec:introduction}
	
		\begin{figure}[t]
		\centering
		\includegraphics[width=0.95\linewidth]{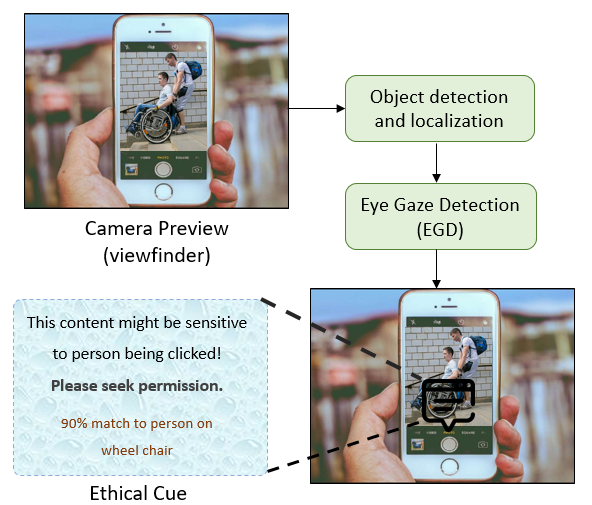}
		\caption{Use of PrivPAS to determine persons with disability in viewfinder image in order to display cues to photographer when subject consent is potentially absent.}
		\label{fig:Pipeline}
	\end{figure}
	In 1890, the United States jurists, Samuel D. Warren and Louis Brandeis authored ``The Right to Privacy'', an article in which they argued for the ``right to be let alone'', using that phrase as a definition of privacy \cite{privacy_definition}. According to Thomas Scanlon, the first element of a theory of privacy should be ``a characterization of the special interest we have in being able to be free from certain kinds of intrusions'' \cite{10.2307/2265077}. Privacy of personally identifiable information (PII) and other user data is one of the major concerns that permeate recent technological developments and associated regulations \cite{pelteret2016review}. Whether it is GDPR in EU or Japan's Act on Protection of Personal Information, nations are bringing legislations and regulations to uphold users' privacy. In major corporate events, like Apple Worldwide Developers Conference (WWDC), Google I/O, Samsung Developer Conference (SDC), there are dedicated sessions explaining how their teams are improving and managing user privacy. At the same time, there is increased attention to developing interfaces and software specifically targeted at enabling people with disabilities. While these indicate that both privacy and accessibility are gaining media and consumer attention, there is yet a lack of focused attempt to mitigate the privacy concerns of people with disabilities although they comprise around 15\% of the world population \cite{who_disability_report}.

	The widespread adoption of online social networks (OSNs) and proliferation of smartphone cameras has democratized visual content creation and consumption, and many do not feel threatened by their photographs appearing on social media \cite{trepte2011privacy}. However, the same technology that makes it easy to share personal details has also led to what Moor \cite{moor1997towards} refers to as ``greased information'' -- data that moves like lightning and is difficult to hold on to. Moor further observes that ``once information is captured electronically for whatever purpose, it is greased and ready to go for any purpose''. Very often, the consent of people in the background of an image is completely overlooked. It is only when the primary subject's consent to being photographed is under question, does the real concern arise. This aggravates in the case of people with disabilities who may feel uncomfortable in raising resentment even when they notice it. Sometimes the images are intentionally collected or clicked to educate people about disability or to inspire others, but this is often regarded to be embarrassing and demoralizing for people with disabilities. ``Inspiration Porn'' is an informal term, coined by the late Australian disability activist Stella Young, for a loose genre of media depictions of people with disabilities \cite{pulrang2019avoid}. This motivates us to encourage ethical and informed decision-making for digital photographers so that they may explicitly seek consent before photographing unaware subjects with disabilities. We aim to explore a solution that may be used to display cues to a photographer based on real-time content in the camera viewfinder as shown in Fig.~\ref{fig:Pipeline}.
	
	In the current work, we present a novel on-device real time Privacy-Preserving AI System (PrivPAS), trained on a custom dataset of annotated images. The core contributions of our work may be summarized as follows:
	
	\begin{enumerate}
		\item We propose a mechanism aimed at an automated real-time detection of human subjects with specific disabilities in a camera viewfinder, indicated by mobility aids and other features.
		\item For identified persons of interest, we present a configuration for determining their awareness of being photographed as a proxy to potential consent.
		\item As the target deployment devices include resource-constrained smartphones, to enable real-time processing of viewfinder images and to ensure no data leaves the device in the process, our solution is designed to be lightweight and efficient as well as entirely on-device.
		\item To further encourage machine learning research on privacy-preserving image data, we benchmark the performance of our object detection to learn certain model parameters, while being trained solely on images, where subject facial information has been deliberately obfuscated. We use the Mobility Aids dataset for the evaluation \cite{vasquez2017deep}\cite{kollmitz2019deep}.
	\end{enumerate}
	
	\section{Related work}\label{sec:relatedWork}
	
	Object detection and localization is a widely studied field in computer vision. Among a large amount of research in this area, we focus only on approaches that integrate both detection and localization in a combined system. Furthermore, special emphasis is laid on approaches for real-time detection on mobile devices.
	
	\subsection{Object detection and localization} \label{sec:CNNsForObjectDetectionAndLocalization}
	
	CNNs \cite{goodfellow2016deep} dominate image detection and classification tasks in computer vision and require relatively little preprocessing compared to other image classification algorithms. CNN learns the filters that in conventional algorithms were hand-engineered thus are independent of prior information and eliminate manual effort in feature design.
	
	Prior to 2012, various techniques were applied to advance the accuracy of object detection on datasets such as PASCAL \cite{everingham2010pascal}. In most of these years, variants of HOG (Histogram of Oriented Gradients) + SVM (Support Vector Machine) \cite{dalal2005histograms} or DPM (Deformable Part Models) \cite{felzenszwalb2009object} were used to define the state-of-art accuracy on these datasets.
	
	In 2012, Alex Net \cite{krizhevsky2012imagenet}, a deep CNN, was used to win the championship in the task of ILSVRC 2012 image classification. Then scholars began to study the application of deep CNN in object detection. They used Alex Net to construct algorithms, such as R-CNN \cite{jiang2017face, girshick2014rich, peng2016multi}, YOLO \cite{redmon2016you}, SSD \cite{redmon2015real}, and others, which resulted in an increased research stream in object detection and localization.
	
	Girshick et al. \cite{girshick2014rich} proposed R-CNN by successfully combining region proposals with CNNs, which improved mean average precision (mAP) by more than 30\%. The next year Girshick \cite{girshick2015fast} named a new algorithm faster R-CNN, which employed spatial pyramid pooling networks. Ren et al. \cite{ren2015faster} introduced a Region Proposal Network (RPN) to overcome few drawbacks of the faster R-CNN. Dai et al. \cite{dai2016r} proposed position-sensitive score maps to address a dilemma between translation-invariance in image classification and translation variance in object detection and successfully executed them 2.5–20 times faster than the F-RCNN counterpart. Lin et al. \cite{lin2017feature} developed a top-down architecture with lateral connections, called Feature Pyramid Network (FPN). All these algorithms successfully solved the problem in object detection. However, there are still defects in accuracy and speed for wireless network object detection applications.
	
	\begin{figure*}[t]
		\centering
		\includegraphics[width=0.7\linewidth]{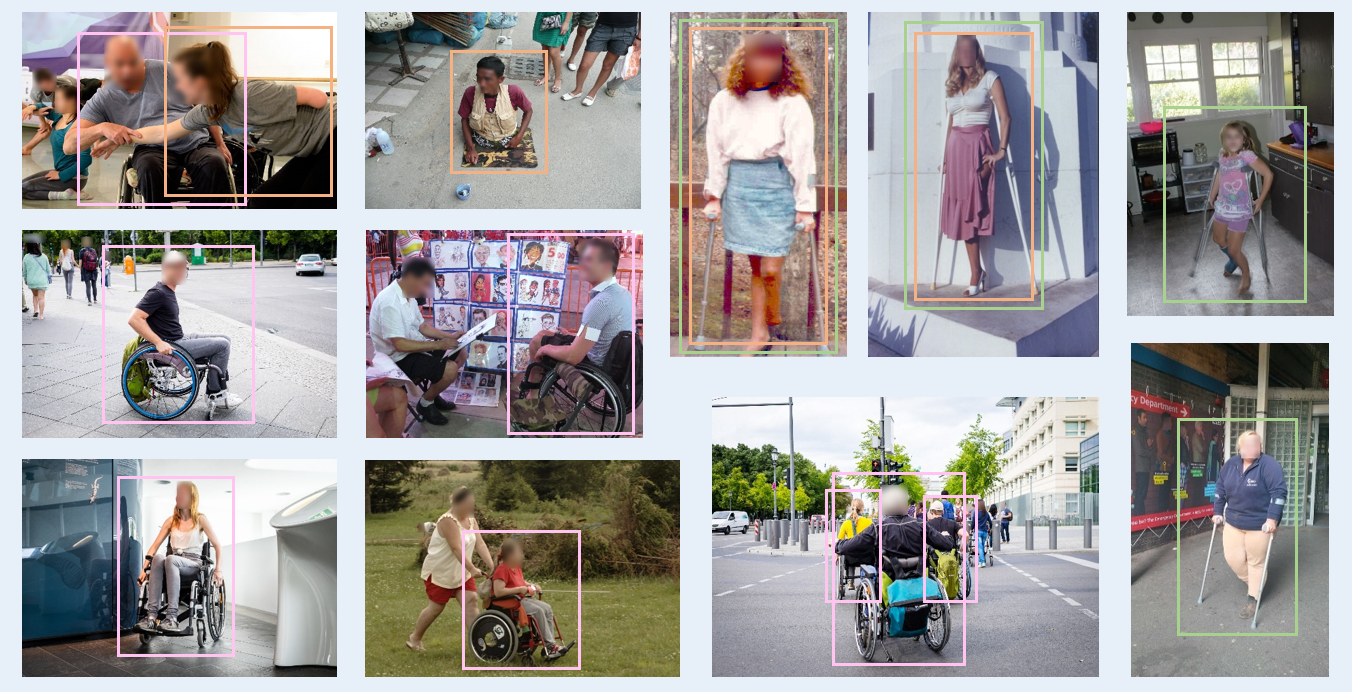}
		\caption{Sample images in dataset labelled with bounded boxes. \textcolor{green}{Green}, \textcolor{pink}{Pink} and \textcolor{orange}{Orange} boxes indicate person using crutches, wheelchair and structurally impaired respectively.}
		\label{fig:sampleImagesInDataset}
	\end{figure*}
	
	To target mobile devices, the primary emphasis is to achieve real-time speed without impacting accuracy levels. To get a better computing speed, Redmon et al. \cite{redmon2016you} proposed a new YOLO algorithm to object detection. It achieved double mAP in real-time detectors. Then Redmon et al. \cite{redmon2017yolo9000} put forward an improved algorithm YOLO V2. Based on YOLO, batch normalization \cite{ioffe2015batch} was added in the algorithm to speed up the training and the algorithm adds anchor boxes and high-resolution classifiers to improve accuracy. Result shows that it runs significantly faster than F-RCNN with ResNet \cite{szegedy2017inception} and SSD \cite{liu2016ssd}. To achieve a high speed and accuracy rate, scholars further optimized the YOLO V2. For instance, Wei et al. \cite{wei2017aerial} used the dimension clustering of object box, classified network pre-training, and other methods. These techniques made the algorithm better adaptive to the location task and object detection. It increased the average accuracy rate to 79.5\%. Redmon et al. \cite{redmon2018yolov3} proposed the third version of the YOLO series, YOLO V3, which improved the algorithm at the accuracy of detection. Researchers have used YOLO versions, and networks inspired by YOLO architecture, like SqueezeDet \cite{8014794}, to target fast, low power on-device object detection.
	
	\subsection{Privacy in Computer Vision}\label{CNNsForPrivacy}
	
	Privacy is important within this domain given the huge range and usage of image recognition algorithms which can potentially expose identifiable and sensitive information. Various research have focused on vision-based pedestrian detection \cite{paul2013human}, human behaviour detection \cite{afsar2015automatic, popoola2012video} and gait recognition \cite{lee2014comprehensive}. Even for privacy, many previous works have targeted use cases such as head inpainting \cite{sun2018natural}. Unfortunately, existing computer vision datasets lack annotated images and videos featuring subjects with disabilities and mobility aids, thus making it difficult to train a system for our task. Furthermore, to the best of our knowledge, there exists no prior work which strives to protect the privacy of people using accessibility or mobility aids.
	
	In contrast to existing work, we propose a novel lightweight on-device network that is capable of decent performance, even when trained on privacy-preserving face-obfuscated image data. Our pipeline identifies and localizes persons with disabilities in viewfinder images in real-time. Unlike generic human subject awareness detection in images, this work focuses on using facial features recognition to determine the awareness of a select group of individuals as a proxy to their potential consent in being photographed, without indulging in person identification.
	
	\section{Dataset: curation and preparation} \label{sec:DatasetCurationAndPreparation}
	
	\subsection{Curation} \label{sec:Curation}
	
	Most AI systems are trained with already existing datasets. In most scenarios, existing datasets fail to capture the complexity of the real world and may lack representation of diverse audiences like people with disabilities. Datasets targetting mobility aids like the Mobility Aids dataset \cite{vasquez2017deep} have a large number of images for the object detection task. But we observe that since these images are curated using video feeds, there is a high image similarity between images captured in consecutive frames and thus, the uniqueness of information within the dataset is very limited. Moreover, there is no existing dataset that is anonymized and annotated based on any possible sensitive content within images. Especially when these images are captured when the subject is unaware. Hence to address these limitations, new datasets representative of people's physical vulnerabilities are needed.
	
\begin{figure}[t]
		\centering
		\includegraphics[width=\linewidth]{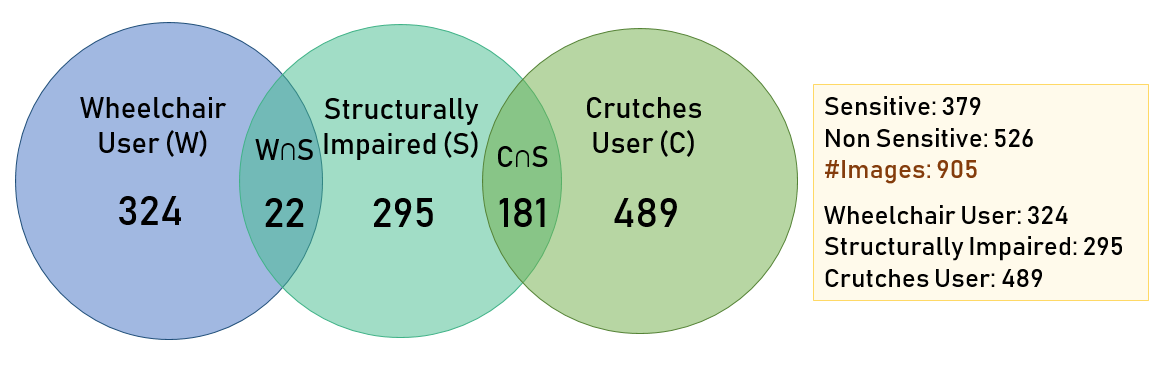}
		\caption{PrivPAS custom dataset statistics}
		\label{fig:PrivPASStatistics}
	\end{figure}

	Therefore, for recognizing any sensitive content concerning the people with disabilities from photos, we curate an image dataset comprising of images of different resolutions, curated from the Flickr open-source and proprietary sources (see Fig.~\ref{fig:PrivPASStatistics}). We use Microsoft VoTT \cite{wadavott} to annotate the dataset with: 
	\begin{enumerate}
		\item Boundary boxes highlighting the individual with disability.
		\item One or more of three accessibility markers -- (a) use of a wheelchair, (b) structural impairment, (c) use of crutches.
		\item Binary label denoting the sensitivity.
	\end{enumerate}

	To reduce the bias in the sensitivity of annotations, the images are labeled by 3 expert annotators of different age groups which ensures diversity as well. The annotated data is then fed to the YOLOv3-Tiny for object detection training on our task which is explained in Section~\ref{sec:model}. Fig.~\ref{fig:sampleImagesInDataset} illustrates sample images in our dataset.
	
	\subsection{Data Augmentation} \label{sec:DataAugmentation}
	The relatively small dataset size is a primary concern from a modeling perspective, as existing works show a correlation between training set size and model performance for object detection tasks. However, this requirement can be addressed using data augmentation and transfer learning.
	
	The purpose of data augmentation is to increase the variability of the input images that can improve the ability of the models to generalize. Photometric distortions and geometric distortions are commonly used data augmentation techniques and help train models with better accuracy. In photometric distortion, we adjust the brightness, contrast, hue, and saturation of an image. For geometric distortion, we apply shifts (horizontal and vertical), flips (horizontal and vertical), and rotation. This is achieved using imgaug \cite{jung2020imgaug} which offers support for bounding boxes (aka rectangles, regions of interest). E.g. if an image is rotated during augmentation, the library can also rotate all bounding boxes on it correspondingly. We use our previously labeled dataset and perform the above-mentioned augmentations along with the bounding boxes to ensure that re-labeling is avoided. To achieve this, we define ranges for each type of augmentation and a random value in this range is chosen for applying the respective augmentation. The ranges used for each type are as follows:
	\begin{figure}[t]
		\centering
		\includegraphics[width=\linewidth]{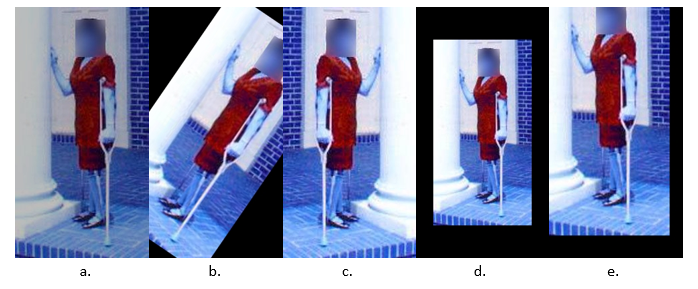}
		\caption{Data augmentation techniques used on curated dataset: a)Brightness adjustment b)Rotation c)Flipping d)Scaling e)Translation.}
		\label{fig:dataAugmentationTechniquesUsed}
	\end{figure}
	
	\begin{enumerate}
		\item Rotation: 0 to 90 
		\item Brightness: 0.2 to 1
		\item Scaling: 0.5 to 1 
		\item Translation: X: -0.2 to 0.3 and Y: -0.1 to 0.3 
		
	\end{enumerate}
	
	The pipeline is set up such that each image is augmented 10 times and thus increases the training set by 10 folds. Fig.~\ref{fig:dataAugmentationTechniquesUsed} shows the output of different augmentations applied to a sample image.
	
	\begin{figure}[b]
		\centering
		\includegraphics[width=\linewidth]{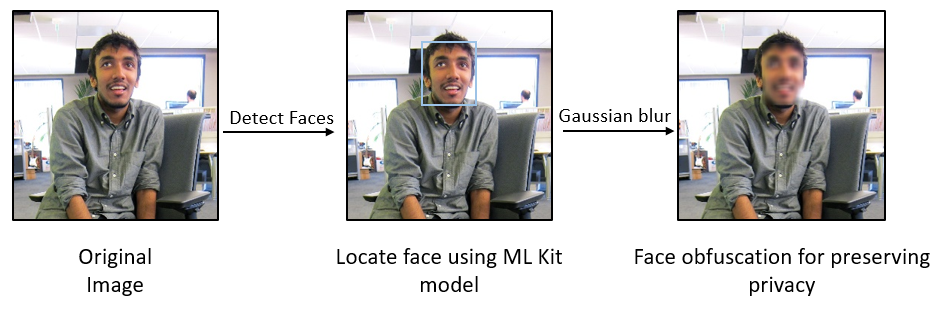}
		\caption{Face obfuscation stages}
		\label{fig:faceObfuscationStages}
	\end{figure}

	\subsection{Data Anonymization} \label{sec:dataAnonymization}
	Faces represent a general and ubiquitous type of private information. Therefore, face obfuscation is widely used for privacy protection in visual datasets. For our proposal to be useful in a context where private user data is involved (for instance, if training data is volunteered by users), we aim to determine the capability of our model to train only on face anonymized dataset and still perform with significant accuracy. We use Gaussian blurring for face de-identification. We use the state-of-the-art ML Kit Face Detection model to detect face contour and generate a bounding box for each face. Suppose there are $m$ face bounding boxes annotated on Image $I$, the bounding box coordinates can be represented using diagonal coordinates:
	\begin{center}
		\begin{equation}
		\label{eq:eq1}
		B_I  = {(x_1^{i},y_1^{i},x_2^{i},y_2^{i} )}_{i=1}^{m}
		\end{equation}
	\end{center}

	To ensure that the entire face is covered by the bounding box, we enlarge each rectangle to be:
	\begin{center}
		\begin{equation}
		\label{eq:eq2}
		B_I^{'}=\biggl\{x_1^{i}-\frac{x_1^{i}}{5}, y_1^{i}-\frac{y_1^{i}}{5}, x_2^{i}+\frac{x_2^{i}}{5}, y_2^{i}+\frac{y_2^{i}}{5} \biggl\}_{i=1}^{m}
		\end{equation}
	\end{center}
	
	A crop of the bounding box is then extracted and we apply Gaussian blurring using the OpenCV library. The Gaussian blurring algorithm scans over each pixel of the cropped image. Each pixel’s new value is essentially computed as the weighted average of the pixels surrounding it. The Gaussian filter in two dimensions is defined as:
	\begin{center}
		\begin{equation}
		\label{eq:eq3}
		G(x,y)=\frac{1}{2\pi\sigma^2} e^{-\frac{x^2+y^2}{2\sigma^2}}
		\end{equation}
	\end{center}
	
	where x and y are the distances from the origin in the x-axis and y-axis respectively, and $\sigma$ is the standard deviation of the Gaussian distribution. We set kernel size as 25 pixels and standard deviation value of kernel as 30 pixels. The values are verified experimentally to effectively remove all identifying facial features. Fig.~\ref{fig:faceObfuscationStages} illustrates the sample results of the entire procedure.

	\section{Model} \label{sec:model}
	
	The pipeline we propose consists of object detection followed by eye-gaze detection as shown in Fig.~\ref{fig:Pipeline}. The details of each component are as follows.
	
	\subsection{Object detection for accessibility markers} \label{sec:objectDetection}
	
	\begin{table*}[t]
		\caption{Model Performance Evaluation}
		\label{table:results}
		\centering
		\resizebox{\textwidth}{!}{\begin{tabular}{ c c c c c c c }
				\toprule
				\multicolumn{1}{c}{\textbf{Model}} &
				\multicolumn{1}{c}{\textbf{Is Tiny}}&
				\multicolumn{1}{c}{\textbf{Data Augmented}}&
				\multicolumn{1}{c}{\textbf{Face Anonymized}}&
				\multicolumn{1}{c}{\textbf{Size (Quantized))}} &
				\multicolumn{1}{c}{\textbf{Validation YOLO Loss}} &
				\multicolumn{1}{c}{\textbf{Validation mAP}}  \\
				\midrule
				
				M1 & No & No  & No & 60.5MB  & 12.997 &  89.46\% \\
				M2 & No &No & Yes  & 60.5MB & 13.429 &  88.92\%  \\
				M3 & No &Yes & No  & 60.5MB & 7.731 &  96.28\%  \\
				M4 & No &Yes & Yes  & 60.5MB &  10.826 & 90.69\% \\
				M5 & Yes &No & No  & 8.49MB &  14.140 & 49.16\%	\\
				M6 & Yes &No & Yes & 8.49MB &  14.705 & 43.50\%	\\
				\textbf{M7} & Yes &Yes & No & \textbf{8.49MB} &  \textbf{7.372} & \textbf{89.58\%}	\\
				\rowcolor{Gray}
				\textbf{M\_Final} & Yes &Yes & Yes & \textbf{8.49MB} &  \textbf{7.809} & \textbf{89.52\%}	\\
				\bottomrule
		\end{tabular}}
	\end{table*}
	Object Detection is a popular task that involves both localization of the required element(s) within an image as well as correctly classifying the element to one of the pre-decided categories. Since our task involves detecting sensitive content concerning the people with disabilities , we first require to identify and locate the presence of these individuals.
	
	For the	accessibility markers detection, we use the approach of one stage detector YOLO (You Only Look Once). It is a deep learning approach for fast object detection. This mainly consists of a deep convolutional network called DarkNet which converts the input to a grid of cells for prediction. For every cell, a bounding box is predicted using five parameters: the coordinates of the center of the target within the box, the height and the width of the box, and the confidence score. The final prediction is based on consolidating and post-processing these candidates.
	
	Following the baseline, there are many variations of this approach. One of the advanced variants is YOLOv3\cite{girshick2014rich, peng2016multi, redmon2015real}. This uses DarkNet 53 containing mainly 3x3 and 1x1 filters along with bypass links. Logistic Regression is used to compute the target scores and thus supports multilabel classification.
	
	Since our requirement is on-device deployment, we explored a smaller version of YOLOv3: YOLOv3-Tiny \cite{redmon2015real}. The processing speed is significantly increased by approximately 442\% as compared to the former variants of YOLO with a marginal reduction in the detection accuracy. For extracting features, convolution layers and max-pooling layers are utilized in the feed-forward arrangement of YOLO v3-Tiny \cite{girshick2014rich, paul2013human, afsar2015automatic}.
	
	Since our dataset is small, to get the best performance, we adopt transfer learning. Here, we use a pre-trained model (on ImageNet \cite{deng2009imagenet}) for the initial stage of training on the custom dataset and in the later stages, we unfreeze the layers and reduce the learning rate. We train our model for a maximum of 51 epochs with early stopping. The YOLO loss consisting of: Co-ordinate loss, Objectiveness loss and Classification loss is used during training. We experimented with 8 model variants as shown in Table~\ref{table:results}. We split the dataset after augmentation into train and validation sets and use the unseen validation set of 905 among a total of 9050 images to evaluate these model variants. The 8 variants include experimenting with anonymized and augmented data as well. We observe that model variants M7 and M\_Final are the most optimal for on-device deployment. However, M\_Final is preferable considering that it performs optimally while being trained on privacy-preserved data.
	. To further enable the on-device deployment of the model, we quantize the trained weights using TFlite. This finally results in a model size of 8.49MB.

	\begin{figure}[b]
		\centering
		\begin{minipage}[b]{0.20\linewidth}
			\includegraphics[width=0.86\linewidth]{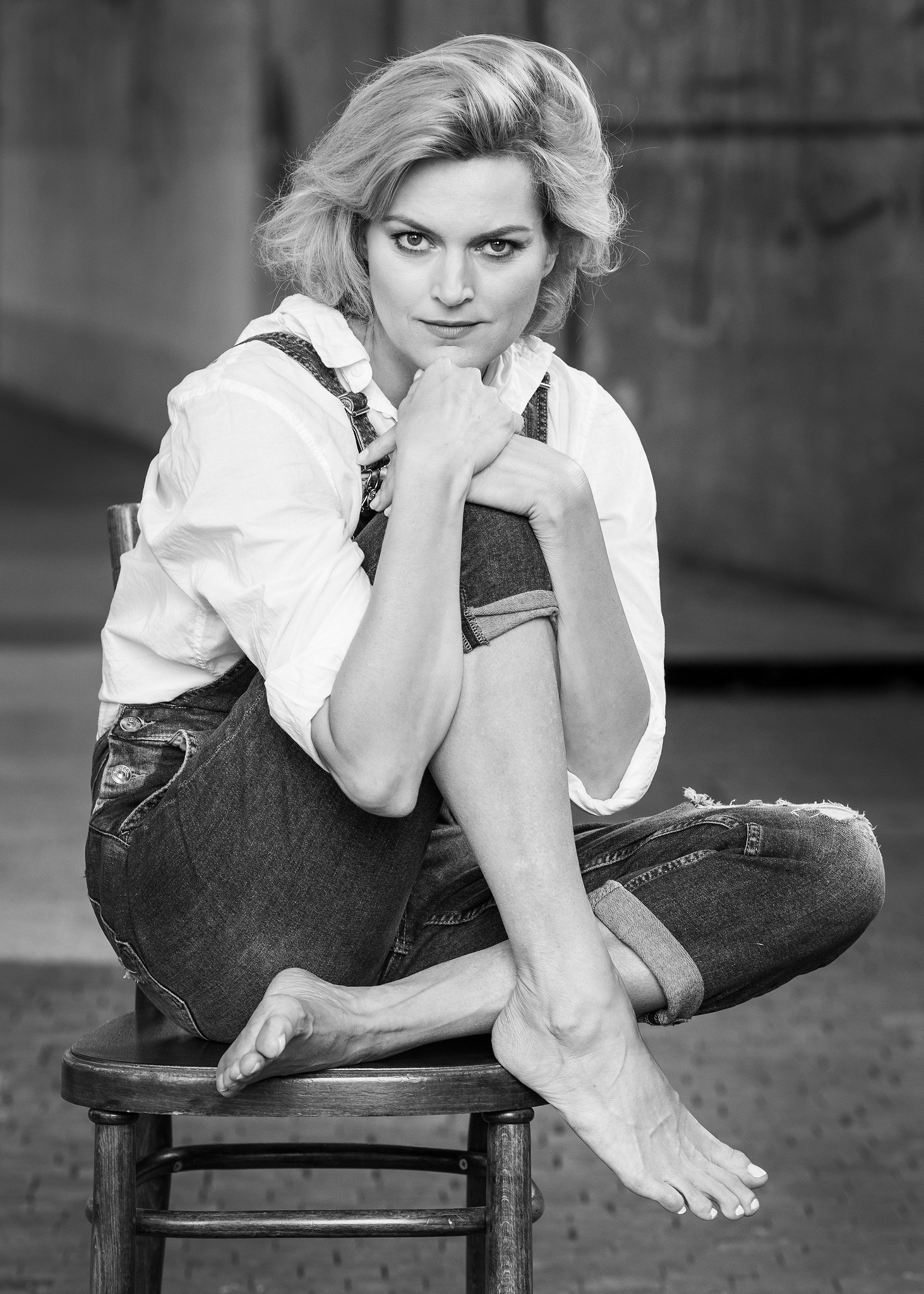}
			\captionsetup{labelformat=empty}
			\caption*{(a) Aware}
			\label{fig:reprImagesForEGD_a}
		\end{minipage}
		\quad
		\begin{minipage}[b]{0.22\linewidth}
			\includegraphics[width=0.86\linewidth]{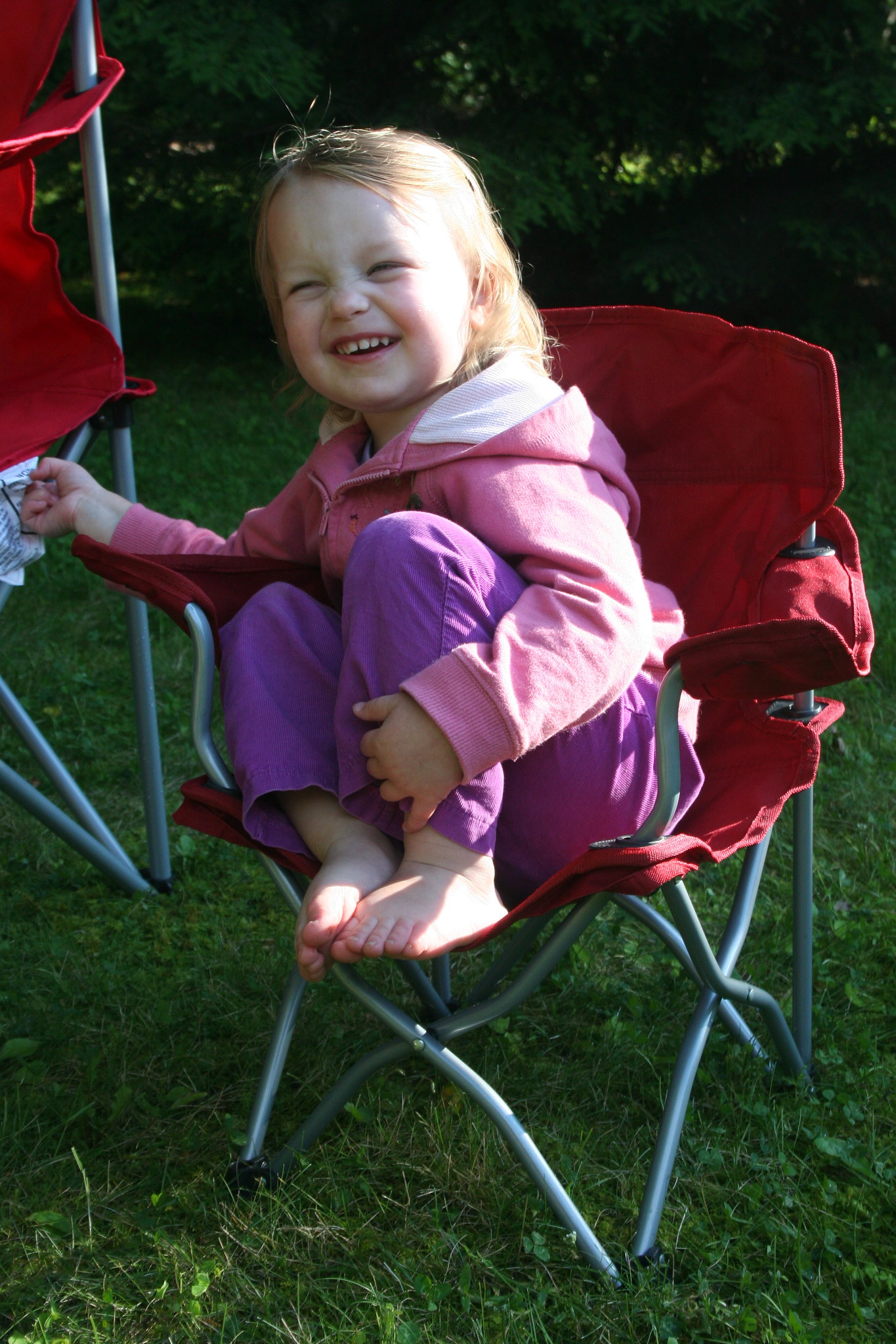}
			\captionsetup{labelformat=empty}
			\caption*{(b) Not aware}
			\label{fig:reprImagesForEGD_b}
		\end{minipage}
		\quad
		\begin{minipage}[b]{0.18\linewidth}
			\includegraphics[width=0.86\linewidth]{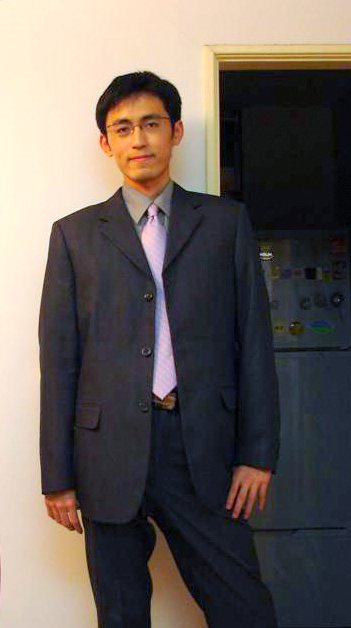}
			\captionsetup{labelformat=empty}
			\caption*{(c) Aware}
			\label{fig:reprImagesForEGD_c}
		\end{minipage}
		\quad
		\begin{minipage}[b]{0.22\linewidth}
			\includegraphics[width=0.86\linewidth]{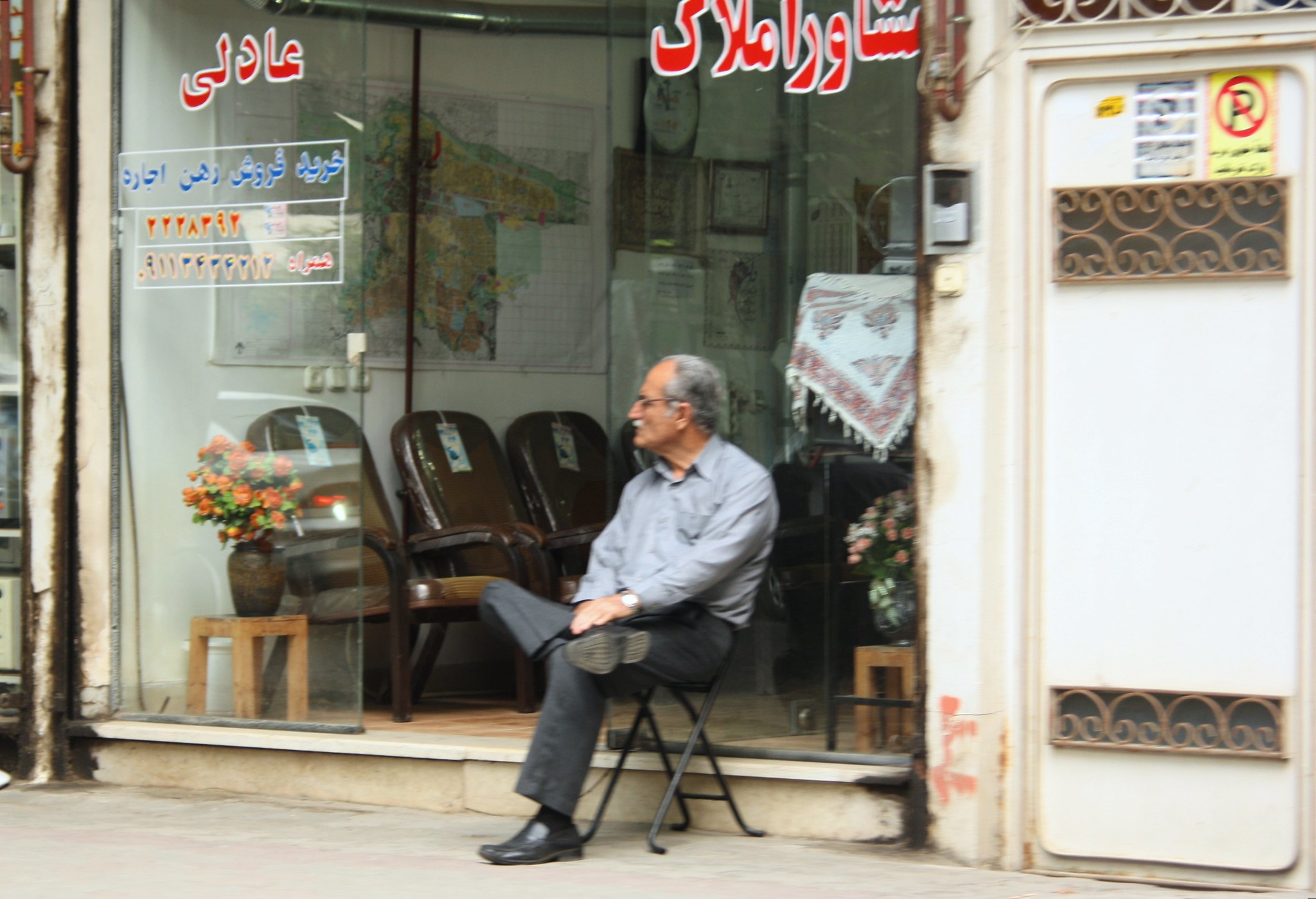}
			\captionsetup{labelformat=empty}
			\caption*{(d) Not aware}
			\label{fig:reprImagesForEGD_d}
		\end{minipage}
		\caption{Representative images for eye gaze detection (with ground truth for POI awareness)}
		\label{fig:reprImagesForEGD}
	\end{figure}

	\begin{figure*}[t]
	\centering
	\includegraphics[width=0.85\linewidth]{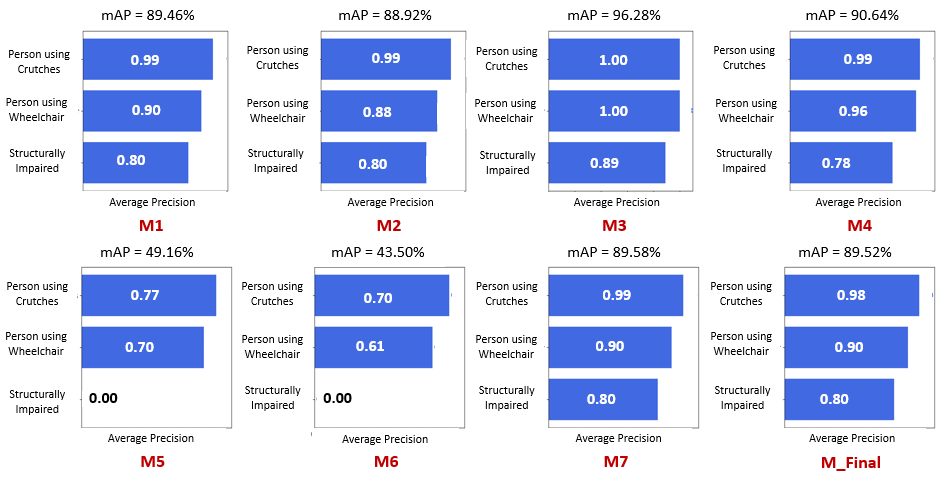}
	\caption{Model Performance on validation set for different variants.}
	\label{fig:validationmAP}
\end{figure*}
	\begin{table*}[b]
		\caption{Eye gaze detection (EGD) on representative images in Fig.~\ref{fig:reprImagesForEGD}}
		\label{tab:eyeGazeStandalone}
		\centering
		\begin{tabular}{ c r c c c r r c c }
			\toprule
			Image & $h_y$   & $P(e_l)$ & $P(e_r)$ & $P(s)$ & $f^r_{POI}$ & $A_{POI}$ & Predicted & Ground Truth \\ \midrule
			(a)   & -1.165  & 0.996    & 0.994    & 0.182  & 0.000       & 0.3397    & Yes       & Yes          \\
			(b)   & -12.069 & 0.195    & 0.823    & 0.997  & -0.047      & -0.6609   & No        & No           \\
			(c)   & -3.582  & 0.997    & 0.999    & 0.012  & 0.000       & 0.4887    & Yes       & Yes          \\
			(d)   & -62.610 & 0.141    & 0.581    & 0.540  & 0.251       & -0.3309   & No        & No           \\ \bottomrule
		\end{tabular}
	\end{table*}

	\subsection{Eye gaze detection (EGD)} \label{sec:eyeGazeDetection}
	
	Eye gaze detection has been the focus of research for various goals, spanning public welfare use cases like traffic accident prevention \cite{naqvi2018deep} and health \cite{jyotsna2018eye, wang2019eye, rupanagudi2019video}, to commercial drivers like television engagement \cite{hernandez2013measuring} and advertisement efficiency \cite{to2021eyes}. In handheld devices like smartphones, user eye gaze tracking has been used for features like power optimization by dimming display brightness when disengaged, and interactive controls like no-touch scrolling \cite{nagamatsu2010mobigaze, bazrafkan2015eye}. However, most of the research literature addresses gaze detection of the smartphone user using the front camera \cite{wood2014eyetab}, where the eyes in question are close to the camera and occupy a significant portion of the viewable range. In contrast, we aim to detect whether a human subject, whose photograph is being captured typically using the rear camera, is aware of the ongoing photography. We also intend to account for candid shots in a private setting, where the subject is conscious of the photography even if they are not directly looking into the camera.
	
	Two major challenges differentiating this task from most existing work are (a) detecting the gaze of eyes, each occupying a very less portion of the image in the viewfinder, and (b) identifying and isolating the facial contours of one person-of-interest (POI) subject from other people who may be present nearby. To tackle these, we leverage the object detection in section \ref{sec:objectDetection} that gives us the bounding box coordinates that enclose the POI. In our on-device deployment, we process a modified image, cropped using the bounding box, using ML Kit Face Detection \cite{ml-kit-face-detection} to derive select facial landmarks and feature contours. We define the relative rotational factor of POI's face as:

	\begin{align}
	f^r_{POI} & =
	\begin{cases}
	1 - \frac{\pi - \left|h_{y}\right| + \left|\tau_r\right|}{\pi - \left|\tau_r\right|}
	& ,
	\left|h_{y}\right| > \left|\tau_r\right|
	\\
	0
	& , \left|h_{y}\right| \le \left|\tau_r\right|
	\end{cases}
	\\
	& =
	\begin{cases}
	\frac{\left|h_{y}\right| - 2\left|\tau_r\right|}{\pi - \left|\tau_r\right|}
	& ,
	\left|h_{y}\right| > \left|\tau_r\right|
	\\
	0
	& , \left|h_{y}\right| \le \left|\tau_r\right|
	\end{cases}
	\end{align}
	
	where, $h_{y}$ denotes the head Euler angle along the $y$-axis, and $\tau_r$ denotes a threshold that is set to 10\textdegree , or $\frac{\pi}{18}$ rad, in our experiments. We use $f^r_{POI}$ to assign an awareness score to the POI as:
	
	\begin{center}
		\begin{equation}
		\begin{multlined}[\columnwidth]
		A_{POI} = w_r * (1 - f^r_{POI}) + w_e * \text{mean}\big(P(e_l), P(e_r)\big) \\ + w_s * P(s) + C
		\end{multlined}
		\end{equation}
	\end{center}

	Here, $P(e_l)$, $P(e_r)$, and $P(s)$ denote the left-eye open probability, right-eye open probability, and smiling probability scores, respectively. The parameters $w_r$, $w_e$, $w_s$ are weights that are learned via logistic regression along with constant bias $C$ to hold the decision boundary for awareness at 0. In our setting, these values were learned as $w_r = -0.07572891$, $w_e = 0.59910001$, $w_s = -0.86601255$, and $C = -0.02311644$.	The results of applying the proposed eye gaze detection method over representative images in Fig.~\ref{fig:reprImagesForEGD} are presented in Table \ref{tab:eyeGazeStandalone}.
	
	We evaluate our eye gaze detection approach as well as the overall usefulness of bounding box coordinates obtained from object detection in determining eye-gaze tracking. Observations from the experiments are presented in the following section.
	
	\begin{figure*}[t]
	\centering
	\includegraphics[width=0.8\linewidth]{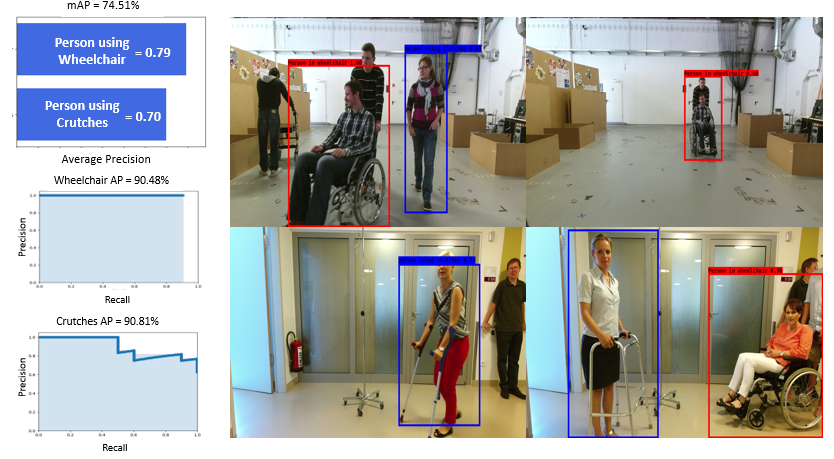}
	\caption{Performance and sample results of model benchmarked on the Mobility Aids dataset.}
	\label{fig:mobilitymAP}
\end{figure*}
	\begin{table*}[t]
		\caption{POI awareness scores using eye gaze detection (EGD) in various settings depict that the use of object detection to identify POI significantly improves the task metrics over na\"ive aggregation of scores from all detected faces}
		\label{tab:eyeGazeAndObjDetection}
		\centering
		\begin{tabular}{ c  c c c c  c c  c }
			\toprule
			Awareness Estimation                  & TP & TN & FP & FN & Precision       & Recall        & F1 Score       \\ \midrule
			EGD $\rightarrow$ \verb|minScoring|   & 113 & 162 & 88  & 137 & 0.562           & 0.452         & 0.501          \\
			EGD $\rightarrow$ \verb|avgScoring|   & 137 & 145 & 105 & 113 & 0.566           & 0.548         & 0.557          \\
			\verb|objDetection| $\rightarrow$ EGD & 189 & 172 & 78  & 61  & \textbf{ 0.708} & \textbf{0.756} & \textbf{0.731} \\ \bottomrule
		\end{tabular}
	\end{table*}
	
	\section{Results and Discussion} \label{sec:results}
	We experiment with our proposed pipeline: PrivPAS on-device using the Samsung A50 smartphone model (6GB RAM, 64GB ROM, Samsung Exynos 7 Octa 9610). Table~\ref{table:results} shows the performance comparison between different variants of our model. We observe that augmentation of images results in better overall model performance. Another notable observation is that the anonymization of data as explained in Section~\ref{sec:dataAnonymization} results in a very small decrease in the model performance which is acceptable considering that the data is now privacy-preserved.

	To measure the model performance of the object detection module, we employ the commonly used Intersection over Union (IoU) and measure the Mean Average Precision (mAP). As shown in Fig.~\ref{fig:validationmAP}, considering model size and performance, M\_Final is the optimal model which has a minimal memory footprint of 8.49MB after quantization and performs with an mAP of 89.52\% on the validation set.
	
	We benchmark our model on the Mobility Aids dataset. Since this dataset does not contain structurally impaired data, we evaluate for the remaining two classes: a person using crutches and a wheelchair. We randomly sample 400 images from the Mobility Aids dataset and measure the performance results. We observe that our model achieves an mAP of 74.51\% as illustrated in Fig.~\ref{fig:mobilitymAP}. The figure also shows the precision-recall curves for both classes and sample model outputs. This shows that our object detection module performs well with a minimal memory footprint.
	
	To evaluate our eye gaze detection approach, we compute the precision, recall, and F1 scores against 500 unseen images. We measure these values in two settings -- (a) with no prior object detection, (b) with object detection described in section \ref{sec:objectDetection}. For scenario (a), where multiple faces are present, we further take two routes to compute the overall score from the individual scores -- minimum and average. These observations are presented in Table~\ref{tab:eyeGazeAndObjDetection}.
	
	Our proposal can help photographers make an ethical choice of seeking consent from subjects for whom such photography may be especially sensitive. It can also enable use cases for automated systems to selectively broadcast recording status, or anonymize people based on disability, or give otherwise preferential treatment to those in need of it. While the current work may not mitigate the possibility of a malicious actor to bypass the ethical cue, it does not add any additional risk to the status quo. In the case of close friends and family members of a photographer, their preference may be recorded using prevalent heuristics like facial recognition, location identifiers, etc. to suppress repetitive prompts in an ethical cue use case.
	
	\section{Conclusion} \label{sec:conclusion}
	
	With the increasing focus on privacy, it is ethical and imperative to cover people from all walks of life. In our research, we observed that: 1) the data concerning people with disabilities is scarce 2) Not much work is available regarding protecting their privacy. In this paper, we present PrivPAS, a lightweight framework with the goal to preserve the privacy of people with disabilities. We start by curating and standardizing a custom image dataset where we explore various augmentation techniques. We systematically compare and analyze a set of lightweight deep learning-based architectures for detecting the accessibility markers.  Then, we show that our proposed model achieves good performance when benchmarked with the Mobility Aids dataset with an mAP of 74.51\% and a size as little as 8.49MB. Further, we explore eye gaze detection and come up with a way to determine the subject's eye gaze from heuristics that are weighted using logistic regression, and achieve an F1 score of 73.1\%. This ascertains the fact that though faces are obfuscated in training images, still good accuracy can be achieved through PrivPAS. We are hopeful that our proposed ethical framework would serve as a starting point to enhance the privacy of individuals with disabilities.

	\bibliographystyle{IEEEtran}
	\bibliography{mybib}

% Generated by IEEEtran.bst, version: 1.14 (2015/08/26)
\begin{thebibliography}{10}
\providecommand{\url}[1]{#1}
\csname url@samestyle\endcsname
\providecommand{\newblock}{\relax}
\providecommand{\bibinfo}[2]{#2}
\providecommand{\BIBentrySTDinterwordspacing}{\spaceskip=0pt\relax}
\providecommand{\BIBentryALTinterwordstretchfactor}{4}
\providecommand{\BIBentryALTinterwordspacing}{\spaceskip=\fontdimen2\font plus
\BIBentryALTinterwordstretchfactor\fontdimen3\font minus
  \fontdimen4\font\relax}
\providecommand{\BIBforeignlanguage}[2]{{%
\expandafter\ifx\csname l@#1\endcsname\relax
\typeout{** WARNING: IEEEtran.bst: No hyphenation pattern has been}%
\typeout{** loaded for the language `#1'. Using the pattern for}%
\typeout{** the default language instead.}%
\else
\language=\csname l@#1\endcsname
\fi
#2}}
\providecommand{\BIBdecl}{\relax}
\BIBdecl

\bibitem{privacy_definition}
\BIBentryALTinterwordspacing
``Privacy,'' {Wikipedia}, accessed on 2021-08-02. [Online]. Available:
  \url{https://en.wikipedia.org/wiki/Privacy}
\BIBentrySTDinterwordspacing

\bibitem{10.2307/2265077}
\BIBentryALTinterwordspacing
J.~Rachels, ``Why privacy is important,'' \emph{Philosophy \& Public Affairs},
  vol.~4, no.~4, pp. 323--333, 1975. [Online]. Available:
  \url{http://www.jstor.org/stable/2265077}
\BIBentrySTDinterwordspacing

\bibitem{pelteret2016review}
M.~Pelteret and J.~Ophoff, ``A review of information privacy and its importance
  to consumers and organizations,'' \emph{Informing Science}, vol.~19, pp.
  277--301, 2016.

\bibitem{who_disability_report}
\BIBentryALTinterwordspacing
``World report on disability,'' {World Health Organization}, 2011. [Online].
  Available: \url{https://www.ncbi.nlm.nih.gov/books/NBK304079/}
\BIBentrySTDinterwordspacing

\bibitem{trepte2011privacy}
S.~Trepte and L.~Reinecke, \emph{Privacy online: Perspectives on privacy and
  self-disclosure in the social web}.\hskip 1em plus 0.5em minus 0.4em\relax
  Springer Science \& Business Media, 2011.

\bibitem{moor1997towards}
J.~H. Moor, ``Towards a theory of privacy in the information age,'' \emph{ACM
  Sigcas Computers and Society}, vol.~27, no.~3, pp. 27--32, 1997.

\bibitem{pulrang2019avoid}
\BIBentryALTinterwordspacing
A.~Pulrang, ``How to avoid ‘inspiration porn.’,'' 2019. [Online].
  Available:
  \url{https://www.forbes.com/sites/andrewpulrang/2019/11/29/how-to-avoid-inspiration-porn/}
\BIBentrySTDinterwordspacing

\bibitem{vasquez2017deep}
A.~Vasquez, M.~Kollmitz, A.~Eitel, and W.~Burgard, ``Deep detection of people
  and their mobility aids for a hospital robot,'' in \emph{2017 European
  Conference on Mobile Robots (ECMR)}.\hskip 1em plus 0.5em minus 0.4em\relax
  IEEE, 2017, pp. 1--7.

\bibitem{kollmitz2019deep}
M.~Kollmitz, A.~Eitel, A.~Vasquez, and W.~Burgard, ``Deep 3d perception of
  people and their mobility aids,'' \emph{Robotics and Autonomous Systems},
  vol. 114, pp. 29--40, 2019.

\bibitem{goodfellow2016deep}
I.~Goodfellow, Y.~Bengio, and A.~Courville, \emph{Deep learning}.\hskip 1em
  plus 0.5em minus 0.4em\relax MIT press, 2016.

\bibitem{everingham2010pascal}
M.~Everingham, L.~Van~Gool, C.~K. Williams, J.~Winn, and A.~Zisserman, ``The
  pascal visual object classes (voc) challenge,'' \emph{International journal
  of computer vision}, vol.~88, no.~2, pp. 303--338, 2010.

\bibitem{dalal2005histograms}
N.~Dalal and B.~Triggs, ``Histograms of oriented gradients for human
  detection,'' in \emph{2005 IEEE computer society conference on computer
  vision and pattern recognition (CVPR'05)}, vol.~1.\hskip 1em plus 0.5em minus
  0.4em\relax Ieee, 2005, pp. 886--893.

\bibitem{felzenszwalb2009object}
P.~F. Felzenszwalb, R.~B. Girshick, D.~McAllester, and D.~Ramanan, ``Object
  detection with discriminatively trained part-based models,'' \emph{IEEE
  transactions on pattern analysis and machine intelligence}, vol.~32, no.~9,
  pp. 1627--1645, 2009.

\bibitem{krizhevsky2012imagenet}
A.~Krizhevsky, I.~Sutskever, and G.~E. Hinton, ``Imagenet classification with
  deep convolutional neural networks,'' \emph{Advances in neural information
  processing systems}, vol.~25, pp. 1097--1105, 2012.

\bibitem{jiang2017face}
H.~Jiang and E.~Learned-Miller, ``Face detection with the faster r-cnn,'' in
  \emph{2017 12th IEEE international conference on automatic face \& gesture
  recognition (FG 2017)}.\hskip 1em plus 0.5em minus 0.4em\relax IEEE, 2017,
  pp. 650--657.

\bibitem{girshick2014rich}
R.~Girshick, J.~Donahue, T.~Darrell, and J.~Malik, ``Rich feature hierarchies
  for accurate object detection and semantic segmentation,'' in
  \emph{Proceedings of the IEEE conference on computer vision and pattern
  recognition}, 2014, pp. 580--587.

\bibitem{peng2016multi}
X.~Peng and C.~Schmid, ``Multi-region two-stream r-cnn for action detection,''
  in \emph{European conference on computer vision}.\hskip 1em plus 0.5em minus
  0.4em\relax Springer, 2016, pp. 744--759.

\bibitem{redmon2016you}
J.~Redmon, S.~Divvala, R.~Girshick, and A.~Farhadi, ``You only look once:
  Unified, real-time object detection,'' in \emph{Proceedings of the IEEE
  conference on computer vision and pattern recognition}, 2016, pp. 779--788.

\bibitem{redmon2015real}
J.~Redmon and A.~Angelova, ``Real-time grasp detection using convolutional
  neural networks,'' in \emph{2015 IEEE International Conference on Robotics
  and Automation (ICRA)}.\hskip 1em plus 0.5em minus 0.4em\relax IEEE, 2015,
  pp. 1316--1322.

\bibitem{girshick2015fast}
R.~Girshick, ``Fast r-cnn,'' in \emph{Proceedings of the IEEE international
  conference on computer vision}, 2015, pp. 1440--1448.

\bibitem{ren2015faster}
S.~Ren, K.~He, R.~Girshick, and J.~Sun, ``Faster r-cnn: Towards real-time
  object detection with region proposal networks,'' \emph{Advances in neural
  information processing systems}, vol.~28, pp. 91--99, 2015.

\bibitem{dai2016r}
J.~Dai, Y.~Li, K.~He, and J.~Sun, ``R-fcn: Object detection via region-based
  fully convolutional networks,'' in \emph{Advances in neural information
  processing systems}, 2016, pp. 379--387.

\bibitem{lin2017feature}
T.-Y. Lin, P.~Doll{\'a}r, R.~Girshick, K.~He, B.~Hariharan, and S.~Belongie,
  ``Feature pyramid networks for object detection,'' in \emph{Proceedings of
  the IEEE conference on computer vision and pattern recognition}, 2017, pp.
  2117--2125.

\bibitem{redmon2017yolo9000}
J.~Redmon and A.~Farhadi, ``Yolo9000: better, faster, stronger,'' in
  \emph{Proceedings of the IEEE conference on computer vision and pattern
  recognition}, 2017, pp. 7263--7271.

\bibitem{ioffe2015batch}
S.~Ioffe and C.~Szegedy, ``Batch normalization: Accelerating deep network
  training by reducing internal covariate shift,'' in \emph{International
  conference on machine learning}.\hskip 1em plus 0.5em minus 0.4em\relax PMLR,
  2015, pp. 448--456.

\bibitem{szegedy2017inception}
C.~Szegedy, S.~Ioffe, V.~Vanhoucke, and A.~A. Alemi, ``Inception-v4,
  inception-resnet and the impact of residual connections on learning,'' in
  \emph{Thirty-first AAAI conference on artificial intelligence}, 2017.

\bibitem{liu2016ssd}
W.~Liu, D.~Anguelov, D.~Erhan, C.~Szegedy, S.~Reed, C.-Y. Fu, and A.~C. Berg,
  ``Ssd: Single shot multibox detector,'' in \emph{European conference on
  computer vision}.\hskip 1em plus 0.5em minus 0.4em\relax Springer, 2016, pp.
  21--37.

\bibitem{wei2017aerial}
Y.~Wei, J.~Quan, and Y.~Hou, ``Aerial image location of unmanned aerial vehicle
  based on yolo v2,'' \emph{Laser \& Optoelectronics Progress}, vol.~54,
  no.~11, p. 111002, 2017.

\bibitem{redmon2018yolov3}
J.~Redmon and A.~Farhadi, ``Yolov3: An incremental improvement,'' \emph{arXiv
  preprint arXiv:1804.02767}, 2018.

\bibitem{8014794}
B.~Wu, A.~Wan, F.~Iandola, P.~H. Jin, and K.~Keutzer, ``Squeezedet: Unified,
  small, low power fully convolutional neural networks for real-time object
  detection for autonomous driving,'' in \emph{2017 IEEE Conference on Computer
  Vision and Pattern Recognition Workshops (CVPRW)}, 2017, pp. 446--454.

\bibitem{paul2013human}
M.~Paul, S.~M. Haque, and S.~Chakraborty, ``Human detection in surveillance
  videos and its applications-a review,'' \emph{EURASIP Journal on Advances in
  Signal Processing}, vol. 2013, no.~1, pp. 1--16, 2013.

\bibitem{afsar2015automatic}
P.~Afsar, P.~Cortez, and H.~Santos, ``Automatic visual detection of human
  behavior: A review from 2000 to 2014,'' \emph{Expert Systems with
  Applications}, vol.~42, no.~20, pp. 6935--6956, 2015.

\bibitem{popoola2012video}
O.~P. Popoola and K.~Wang, ``Video-based abnormal human behavior
  recognition—a review,'' \emph{IEEE Transactions on Systems, Man, and
  Cybernetics, Part C (Applications and Reviews)}, vol.~42, no.~6, pp.
  865--878, 2012.

\bibitem{lee2014comprehensive}
T.~K. Lee, M.~Belkhatir, and S.~Sanei, ``A comprehensive review of past and
  present vision-based techniques for gait recognition,'' \emph{Multimedia
  tools and applications}, vol.~72, no.~3, pp. 2833--2869, 2014.

\bibitem{sun2018natural}
Q.~Sun, L.~Ma, S.~J. Oh, L.~Van~Gool, B.~Schiele, and M.~Fritz, ``Natural and
  effective obfuscation by head inpainting,'' in \emph{Proceedings of the IEEE
  Conference on Computer Vision and Pattern Recognition}, 2018, pp. 5050--5059.

\bibitem{wadavott}
K.~Wada, ``Vott: Visual object tagging tool.''

\bibitem{jung2020imgaug}
A.~B. Jung, K.~Wada, J.~Crall, S.~Tanaka, J.~Graving, S.~Yadav, J.~Banerjee,
  G.~Vecsei, A.~Kraft, J.~Borovec \emph{et~al.}, ``Imgaug,'' \emph{GitHub: San
  Francisco, CA, USA}, 2020.

\bibitem{deng2009imagenet}
J.~Deng, W.~Dong, R.~Socher, L.-J. Li, K.~Li, and L.~Fei-Fei, ``Imagenet: A
  large-scale hierarchical image database,'' in \emph{2009 IEEE conference on
  computer vision and pattern recognition}.\hskip 1em plus 0.5em minus
  0.4em\relax Ieee, 2009, pp. 248--255.

\bibitem{naqvi2018deep}
R.~A. Naqvi, M.~Arsalan, G.~Batchuluun, H.~S. Yoon, and K.~R. Park, ``Deep
  learning-based gaze detection system for automobile drivers using a nir
  camera sensor,'' \emph{Sensors}, vol.~18, no.~2, p. 456, 2018.

\bibitem{jyotsna2018eye}
C.~Jyotsna and J.~Amudha, ``Eye gaze as an indicator for stress level analysis
  in students,'' in \emph{2018 International Conference on Advances in
  Computing, Communications and Informatics (ICACCI)}.\hskip 1em plus 0.5em
  minus 0.4em\relax IEEE, 2018, pp. 1588--1593.

\bibitem{wang2019eye}
Y.~Wang, R.~Huang, and L.~Guo, ``Eye gaze pattern analysis for fatigue
  detection based on gp-bcnn with esm,'' \emph{Pattern Recognition Letters},
  vol. 123, pp. 61--74, 2019.

\bibitem{rupanagudi2019video}
S.~R. Rupanagudi, V.~G. Bhat, S.~K. Gurikar, S.~P. Koundinya, S.~Kumar,
  R.~Shreyas, S.~Shilpa, N.~Suman, R.~R. Bademi, M.~Koppisetti \emph{et~al.},
  ``A video processing based eye gaze recognition algorithm for wheelchair
  control,'' in \emph{2019 10th International Conference on Dependable Systems,
  Services and Technologies (DESSERT)}.\hskip 1em plus 0.5em minus 0.4em\relax
  IEEE, 2019, pp. 241--247.

\bibitem{hernandez2013measuring}
J.~Hernandez, Z.~Liu, G.~Hulten, D.~DeBarr, K.~Krum, and Z.~Zhang, ``Measuring
  the engagement level of tv viewers,'' in \emph{2013 10th IEEE International
  Conference and Workshops on Automatic Face and Gesture Recognition
  (FG)}.\hskip 1em plus 0.5em minus 0.4em\relax IEEE, 2013, pp. 1--7.

\bibitem{to2021eyes}
R.~N. To and V.~M. Patrick, ``How the eyes connect to the heart: The influence
  of eye gaze direction on advertising effectiveness,'' \emph{Journal of
  Consumer Research}, vol.~48, no.~1, pp. 123--146, 2021.

\bibitem{nagamatsu2010mobigaze}
T.~Nagamatsu, M.~Yamamoto, and H.~Sato, ``Mobigaze: Development of a gaze
  interface for handheld mobile devices,'' in \emph{CHI'10 Extended Abstracts
  on Human Factors in Computing Systems}, 2010, pp. 3349--3354.

\bibitem{bazrafkan2015eye}
S.~Bazrafkan, A.~Kar, and C.~Costache, ``Eye gaze for consumer electronics:
  Controlling and commanding intelligent systems.'' \emph{IEEE Consumer
  Electronics Magazine}, vol.~4, no.~4, pp. 65--71, 2015.

\bibitem{wood2014eyetab}
E.~Wood and A.~Bulling, ``Eyetab: Model-based gaze estimation on unmodified
  tablet computers,'' in \emph{Proceedings of the Symposium on Eye Tracking
  Research and Applications}, 2014, pp. 207--210.

\bibitem{ml-kit-face-detection}
\BIBentryALTinterwordspacing
``Face detection | ml kit,'' {Google Developers}, accessed on 2021-07-26.
  [Online]. Available:
  \url{https://developers.google.com/ml-kit/vision/face-detection}
\BIBentrySTDinterwordspacing

\end{thebibliography}
	
\end{document}